\documentclass{article}

\PassOptionsToPackage{numbers, compress}{natbib}

 \usepackage[dblblindworkshop, final]{neurips_2025}
\workshoptitle{New Perspectives in Graph Machine Learning
}

\usepackage[utf8]{inputenc} 
\usepackage[T1]{fontenc}    
\usepackage{hyperref}       
\usepackage{url}            
\usepackage{booktabs}       
\usepackage{amsfonts}       
\usepackage{nicefrac}       
\usepackage{microtype}      
\usepackage{xcolor}         

\usepackage{caption}
\usepackage{multirow}
\usepackage{amsmath}
\usepackage{subfiles}
\usepackage{array}
\usepackage{graphicx}
\usepackage{subfigure}
\usepackage{amsthm}
\usepackage{authblk}
\title{Predict Training Data Quality via Its Geometry in Metric Space}

\author{\textbf{Yang Ba}}
\author{\textbf{Mohammad Sadeq Abolhasani}}
\author{\textbf{Rong Pan}}
\affil{School of Computing and Augmented Intelligence, Arizona State University\\
\texttt{\{yangba, mabolhas, Rong.Pan\}@asu.edu}}

\begin{document}

\maketitle

\begin{abstract}

High-quality training data is the foundation of machine learning and artificial intelligence, shaping how models learn and perform. Although much is known about what types of data are effective for training, the impact of the data's geometric structure on model performance remains largely underexplored. We propose that both the richness of representation and the elimination of redundancy within training data critically influence learning outcomes. To investigate this, we employ persistent homology to extract topological features from data within a metric space, thereby offering a principled way to quantify diversity beyond entropy-based measures. Our findings highlight persistent homology as a powerful tool for analyzing and enhancing the training data that drives AI systems.







\end{abstract}

\section{Introduction}

Data serves as the cornerstone of the artificial intelligence (AI) revolution, with its quality directly shaping the performance and reliability of AI models.  
Specifically, the features, patterns, and information embedded in data determine how effectively models can learn \cite{bishop2006pattern,lecun2015deep}. Yet not all data carries equal value. Coverage and diversity, in particular, play a pivotal role in model performance, influencing the generalization, fairness, and robustness of AI systems \cite{rolf2021representation,clemmensen2022data, pmlr-v258-kim25f}. This raises an important question: which specific properties of data make it most valuable for model training, and how can we systematically construct high-quality datasets that embody them?

\begin{figure}[h]
    \centering
    \includegraphics[width=0.5\linewidth]{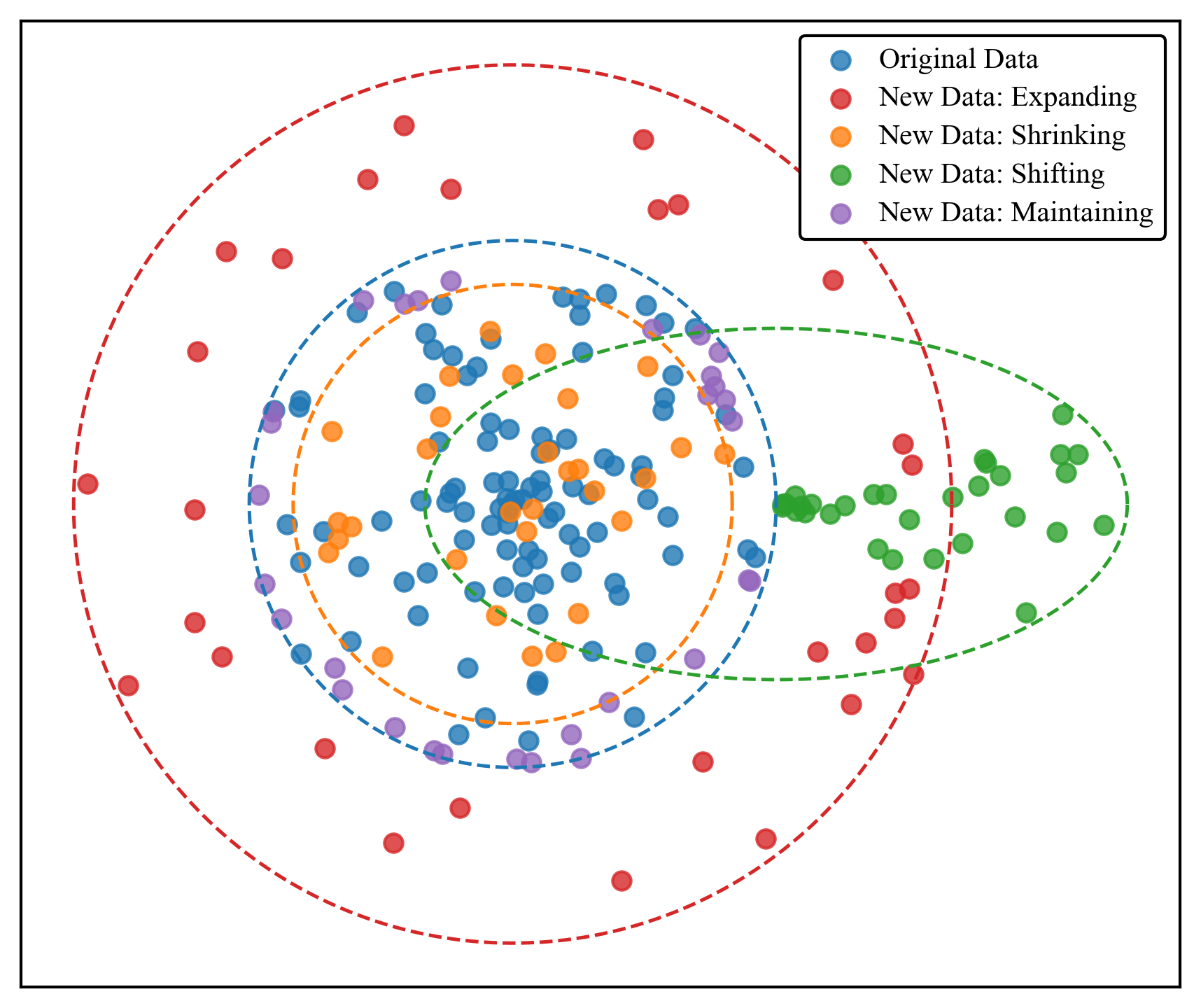}
    \caption{There are four scenarios for adding new data points to augment the current dataset (blue). However, without prior knowledge of the unseen data, clear guidance on which approach is most effective is lacking.}
    \label{fig:intro}
\end{figure}

A recent study \citep{ba2024does} has shown a strong link between training data diversity and model performance, demonstrating that greater diversity improves both in-distribution (ID) and out-of-distribution (OOD) generalization. The importance of diversity has long been recognized in machine learning, with data augmentation serving as a common strategy for introducing data variability \cite{NEURIPS2021_fb4c4860, shorten2019survey, zhang2017mixup}. By exposing models to a broader range of scenarios and feature variations, diverse training data reduces overfitting and enhances generalization to unseen cases. Building on this perspective, we characterize data quality in terms of its diversity and hypothesize that higher-quality, more diverse data can lead to better model performance. To quantify such diversity, metrics such as the Vendi Score \cite{dan2023vendi} have been introduced. This entropy-based approach is inspired by the concept of "community diversity" in ecology and biology \cite{daly2018ecological, leinster2021entropy}. However, excessive diversity can also be detrimental, potentially introducing distribution shifts that degrade performance. This raises several important questions: 
\emph{Aside from class balance, if all data points are equally important, will more data always be beneficial?} \emph{When augmenting a dataset, what types of data are more valuable to add?} These questions motivate our focus on the geometry of data and its role in shaping model performance. Considering a dataset embedded in a metric space (Figure \ref{fig:intro}) that we wish to augment, four augmentation strategies are possible -- shrinking, expanding, maintaining, or shifting the scope of the data. The central challenge, then, is to determine which of these strategies most effectively enhances generalization.

The biology-inspired, entropy-based diversity measures can capture how evenly data are distributed from a purely distributional perspective. Building on the fundamental connection between persistent homology (PH) and agglomerative hierarchical clustering \cite{murtagh2012algorithms}, we propose a PH-based diversity measure that extends this idea to capture the topological features of data. First, we show that the PH-based measure satisfies the axiomatic definition of diversity. Next, we demonstrate how the geometry of training data influences model performance through a transfer learning framework for classification tasks. Finally, we provide practical guidance for training dataset augmentation and data point selection informed by these findings.

Our contributions are summarized as follows: 
\begin{itemize}
    \item We show that persistent homology, as a diversity measure, can capture richer structural information than conventional distribution- or entropy-based metrics.
    
    \item We develop multiple PH–based diversity measures that can quantify data quality and reveal their connection to model performance, highlighting the value of higher-order data geometric features (e.g., \(H_1\)) playing a key role in capturing meaningful structural patterns.  

\end{itemize}

Our study offers a deeper understanding of the role of data geometry in the generalizability of AI models, paving the way for its integration into workflows for data augmentation, data selection, and synthetic data generation.





\section{Background}
\subsection{Persistent Homology in Metric Space}

Persistent Homology (PH) \cite{edelsbrunner2002topological, edelsbrunner2008persistent} is a central tool in Topological Data Analysis (TDA) for uncovering the underlying shape of data, which is typically represented as a point cloud. This technique constructs a sequence of geometric objects, called simplicial complexes, over continuously expanding scales to connect nearby data points. By applying an algebraic tool, called \emph{homology}, PH tracks the birth and death of topological features such as connected components, loops, and voids as the scale grows. The result is a multi-scale summary, commonly visualized as a barcode or persistence diagram, which distinguishes significant, long-lived features from noise. The Stability Theorem ensures that these summaries are robust to small perturbations in the data, making PH a reliable tool for extracting meaningful data structure information. While PH has been widely applied in various domains \cite{zhao2019learning,hiraoka2016hierarchical, pun2022persistent}, its potential as a direct measure of data diversity remains largely unexplored.

\subsection{Diversity Measurement}

Several metrics have been developed to quantify the diversity of a dataset. Among them, Vendi Score (VS) \citep{dan2023vendi} is the most prominent one and is often used in various data augmentation tasks. This score, derived from a set of samples and their pairwise similarity functions, quantifies the similarities among the data in a dataset. 
Mathematically, VS is given by the exponential of the Shannon entropy, which is obtained from the eigenvalues of the scaled similarity matrix \( X^\top X \):
\[
VS 
= \exp \left( - \sum_{i=1}^{n} \lambda_i \log \lambda_i \right)
\]
where \( \lambda_i \) are the eigenvalues of scaled \( X^\top X \).  
Another work \cite{limbeck2024metric} introduces several magnitude-based diversity measures that quantify the effective size of a space across scales. The magnitude function, $Mag_X(t)$, captures data diversity from local to global scopes. Specifically, the authors proposed two metrics, MAGAREA and MAGDIFF, which provide robust measures of intrinsic diversity and enable meaningful comparisons between datasets, particularly for detecting mode collapse.

\section{Methodology}

\subsection{PH-Based Diversity Measure}

We define PH-based diversity measures by using persistent homology (PH) lifetimes derived from a Vietoris–Rips complex constructed on the pairwise distance matrix of the dataset. Let $X = \{x_1, \dots, x_n\} \subset \mathbb{R}^d$ be the dataset and $D$ denote its pairwise distance matrix. The choice of distance metric depends on the application, with Euclidean distance and cosine distance being common options. Intuitively, longer lifetimes correspond to more persistent topological structures, reflecting the importance of the underlying data geometry; thus, these PH lifetimes provide a principled way to quantify diversity beyond conventional distributional measures.

To start, we construct a Vietoris–Rips filtration $\{\text{VR}_\epsilon(D)\}_{\epsilon \geq 0}$ \cite{vietoris1927hoheren}, where simplices are included whenever the corresponding pairwise distances in $D$ are less than or equal to $\epsilon$. Persistent homology yields a set of intervals \cite{ghrist2008barcodes}:
$$
\mathcal{B}_k = \{ (b_i, d_i) \}_{i=1}^{m_k}, \quad k = 0,1,2,\dots
$$
where $\mathcal{B}_k$ denotes the set of persistence intervals in homological dimension $k$, and $m_k = |\mathcal{B}_k|$ is the total number of such intervals. Each interval corresponds to a $k$-dimensional topological feature (e.g., connected components when $k=0$, loops when $k=1$, voids when $k=2$, etc.), with birth scale $b_i$ and death scale $d_i$. The persistence length (or lifetime) of the $i-th$ feature in homological dimension $k$ is then defined as
$$
l_i = d_i - b_i.
$$
In this work, we restrict our analysis to 0-dimensional features ($H_0$, connected components) and 1-dimensional features ($H_1$, loops). To summarize diversity, we consider persistence-based analogues of entropy and Hill numbers. Define the normalized persistence weights as 
\[
p_i = \frac{l_i}{L},
\quad \text{where} \quad
L = \sum_{i=1}^{m_k} l_i.
\]
The Rényi persistence entropy of order \(q \ge 0,\, q \neq 1\) is then defined as
\[
\mathrm{PE}_k^{(q)} 
= \frac{1}{1-q} \log\!\Bigl(\sum_{i=1}^{m_k} p_i^{\,q}\Bigr).
\]
As \(q \to 1\), this reduces to the Shannon persistence entropy \cite{merelli2015topological}:
\[
\mathrm{PE}_k^{(1)} 
= - \sum_{i=1}^{m_k} p_i \,\log p_i,
\]
Finally, the corresponding PH-based Hill numbers ($\mathrm{PEH}$) can be expressed as the exponential of the Rényi persistence entropy, analogous to Hill numbers in ecology :
\[
\mathrm{PEH}_k^{q}(X) = \exp\bigl(\mathrm{PE}_k^{(q)}\bigr).
\]
PEH quantifies the effective number of topologically significant features in the dataset. By varying \(q\), one can emphasize either rare or dominant features.

Unlike entropy-based measures, which quantify distributional richness in terms of eigenvalue spectra, PH-based diversity 
quantifies the stability of the corresponding topological structure across scales. Longer persistence lengths correspond to more significant and robust features, thereby providing a natural foundation for defining diversity measures that reflect thegeometric and topological variability in the data.

\subsection{Axiomatic Definition of Diversity}

Entropy-based measures, such as magnitude function and Vendi Score, are strongly connected to the notion of diversity and are widely used in ecological studies due to their computational efficiency and adherence to core diversity axioms \cite{leinster2012measuring, limbeck2024metric}. The key principles are: 
\begin{itemize}
    \item \textbf{Effective size: } In a dataset of fixed size, diversity increases when data points are well-separated and decreases as they cluster, reaching a maximum when all points are distinct and a minimum when all are identical.
    \item \textbf{Twin property: } Adding a duplicate observation leaves diversity unchanged.
    \item \textbf{Multi-scale: } Diversity is evaluated across multiple scales of similarity, capturing both local and global structure in the data manifold.
    \item \textbf{Symmetry: } Diversity is invariant to the order of data points, exhibiting permutation invariance.
\end{itemize}

The proposed PH-based diversity measure satisfies these principles. For \emph{Effective size},  when all data points overlap, only one connected component ($H_0$) merges and no loops ($H_1$) form, yielding short lifetimes and near-zero diversity. Conversely, multiple persistent clusters ($H_0$) or robust loops ($H_1$) produce long, varied lifetimes. Summary statistics such as total persistence, mean, and variance of $H_0$ and $H_1$, or the PH-based Hill number capture higher diversity, reflecting the geometric spread of the data. For \emph{twin property}, adding a duplicate point has no effect, as a duplicate $x_n$ of $x_i \in X$ has zero distance to its twin and identical distances to all other points. In the Vietoris–Rips filtration, it immediately merges with $x_i$ and does not create any new feature with non-zero persistence. Consequently, all non-zero persistence intervals, their weights, and the final diversity measure remain unchanged. 
Different homological dimensions capture geometric features across multiple scales through the filtration parameter $\epsilon$, while the Hill-number order $q$ adjusts the emphasis between local and global structures. The computation relies solely on the pairwise distance matrix, ensuring invariance to the ordering of samples. Formal proofs are provided in Appendix~\ref{appedix}.

\subsection{Relationship between Structural Diversity and Model Performance}

\begin{figure}[htb]
    \centering
    \includegraphics[width=0.45\linewidth]{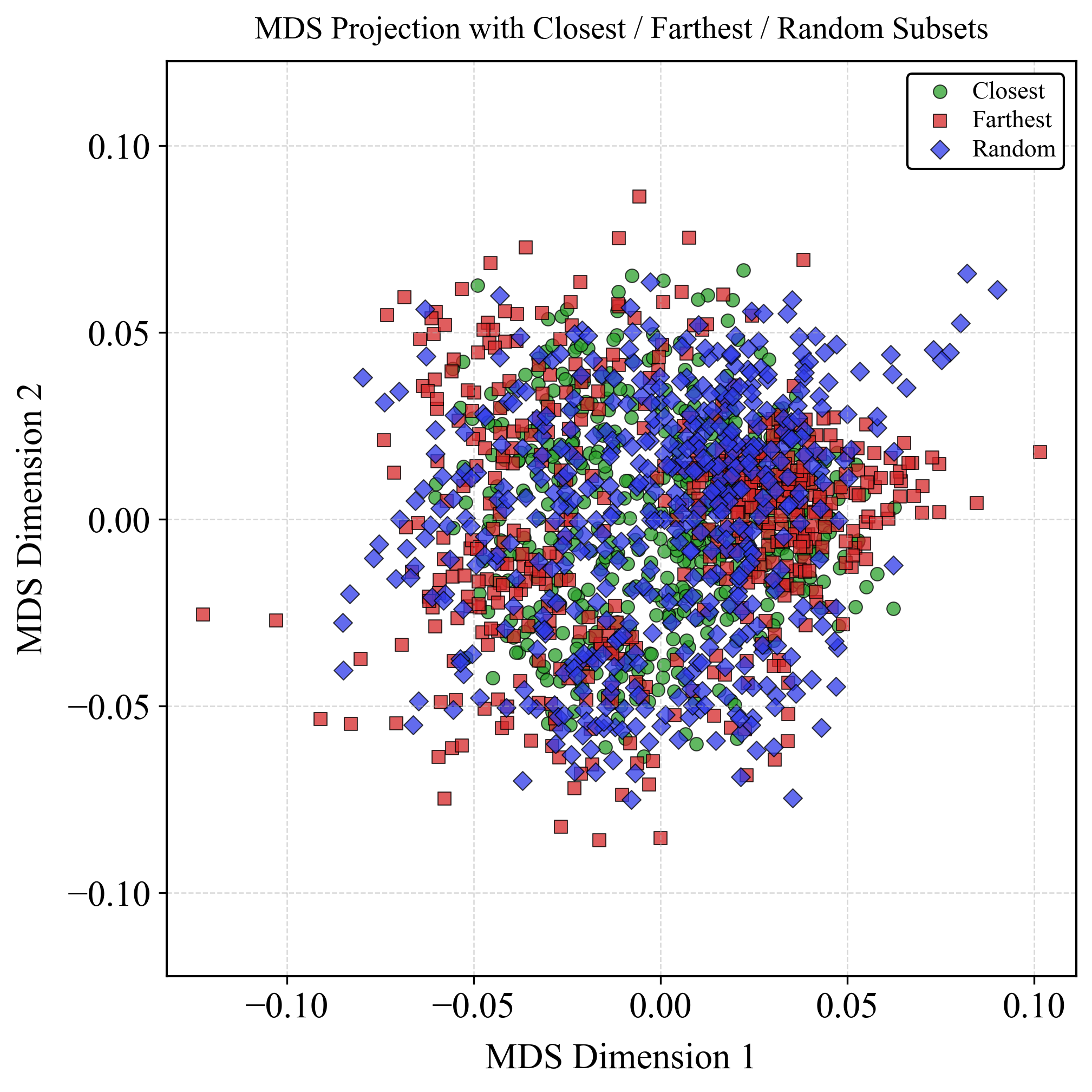}
    \caption{A demonstration of three representative subsets construction for "Medical" dataset}
    \label{fig:mds}
\end{figure}

To investigate the relationship between structural diversity in metric space and model performance, we constructed three balanced subsets (the closest, the farthest, and random) based on pairwise distance matrices. For each sample, the maximum distance from a data point to all other points was computed, and it is used to rank points. The closest subset was drawn from the lower half of this ranking (core samples), while the farthest subset was drawn from the upper half (peripheral samples). In each case, equal numbers of class 0 and class 1 samples were randomly selected to ensure a balanced dataset. As a baseline, the random subset was created by sampling equal numbers of class 0 and class 1 directly from all data points, regardless of their distance rankings. Figure \ref{fig:mds} illustrates an example of how three subsets are distributed in 2D space using multidimensional scaling (MDS) \cite{davison2000multidimensional}. By analyzing the persistent homology summaries of these representative subsets and comparing the performance (accuracy) of models trained upon them, we aim to uncover systematic connections between the geometric structure of data and model generalization.

\section{Experiment \& Analysis}

\subsection{Experiment Setup}
We evaluate our hypothesis on text classification tasks across multiple domains using a transfer learning approach. Specifically, we fine-tune BERT$_{base}$ models \cite{devlin2018bert} by adding a dropout layer and a softmax classifier on top of the pre-trained architecture. Each model is trained for 8 epochs with a learning rate of 1e-6 and a dropout rate of 10\%. 

The evaluation spans several datasets -- the Complaints dataset (TC) \cite{preotiuc2019automatically}, the SUBJectivity dataset (SUBJ) \cite{pang2004sentimental}, SentEval (SE) \cite{hu2004mining}, Arxiv-10 \cite{farhangi2022protoformer}, and Medical \cite{fansi2022ddxplus}. For each dataset, we constructed three subsets—closest, farthest, and random—and ran experiments three times per subset. To ensure consistent experimental conditions, each training set contains 500 samples (250 per class) across all datasets and subsets. Our experiments are limited to text classification with BERT fine-tuning; extending the evaluation to other modalities and larger-scale settings is left for future work.

\subsection{Experiment Result}

\begin{figure}[htb]
    \begin{subfigure}
    \centering
    \includegraphics[width=0.33\textwidth]{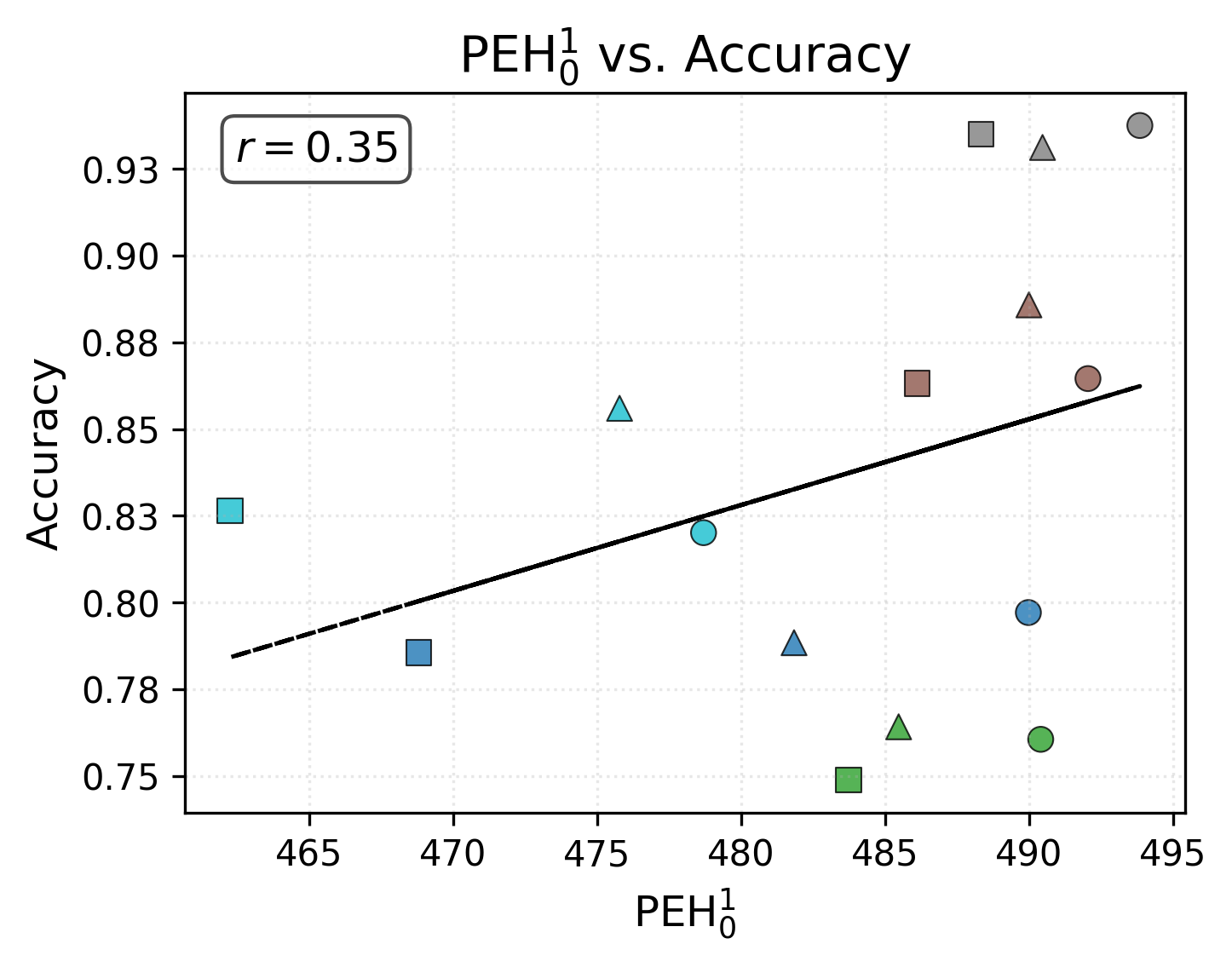}
    \includegraphics[width=0.33\textwidth]{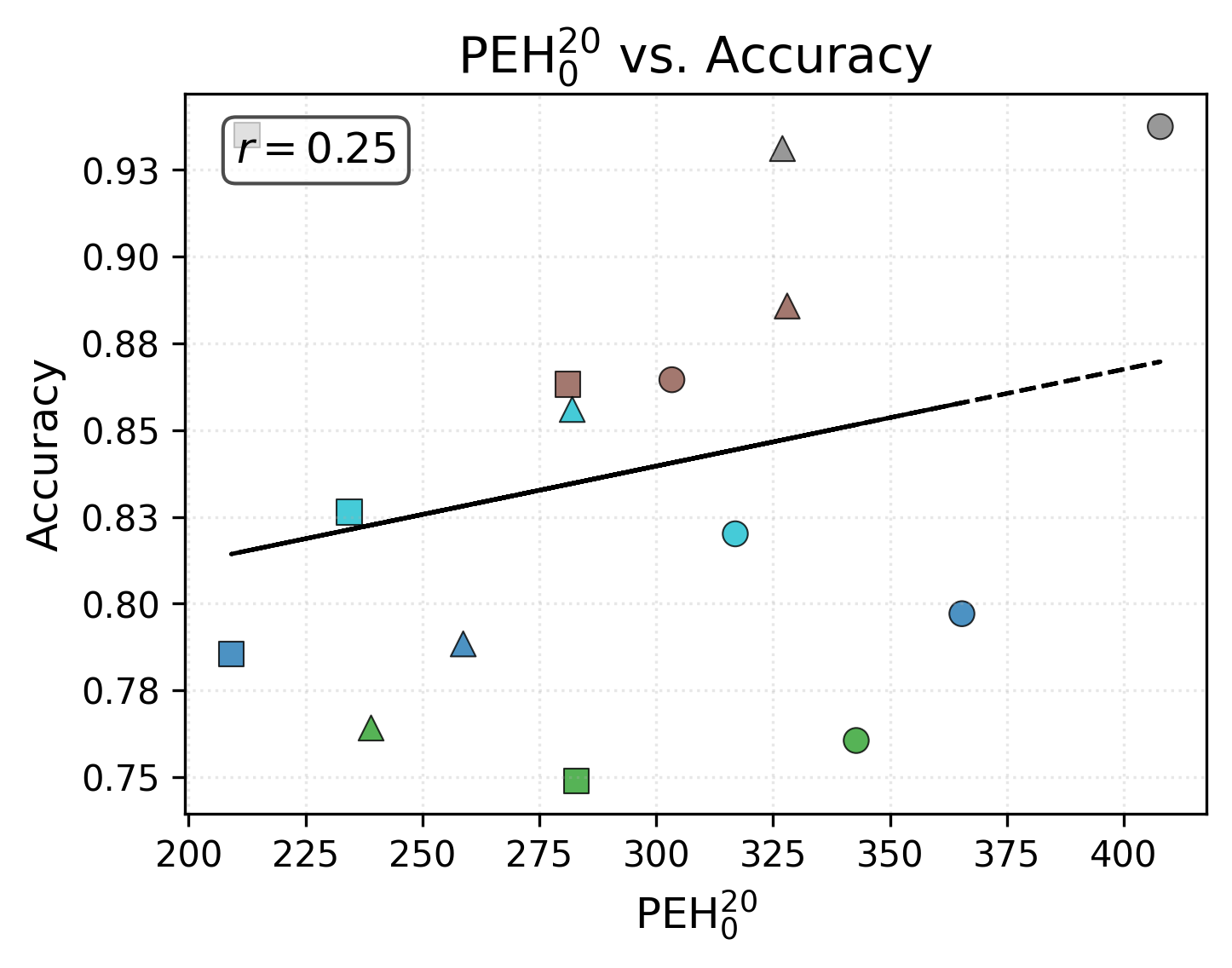}
    \includegraphics[width=0.33\textwidth]{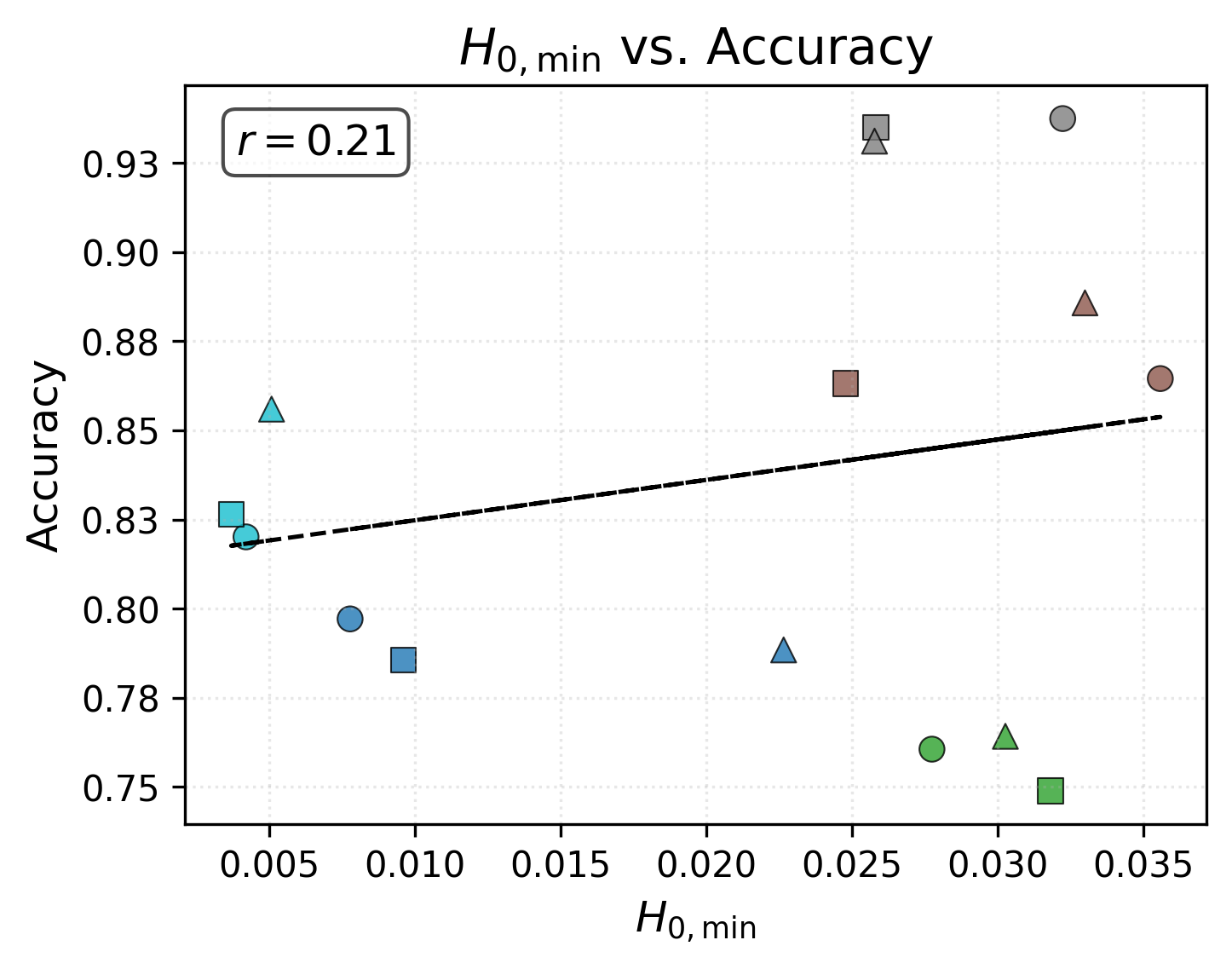}
    \label{fig:h0}
    \end{subfigure}
    \vfill
    \begin{subfigure}
      \centering
    \includegraphics[width=0.33\textwidth]{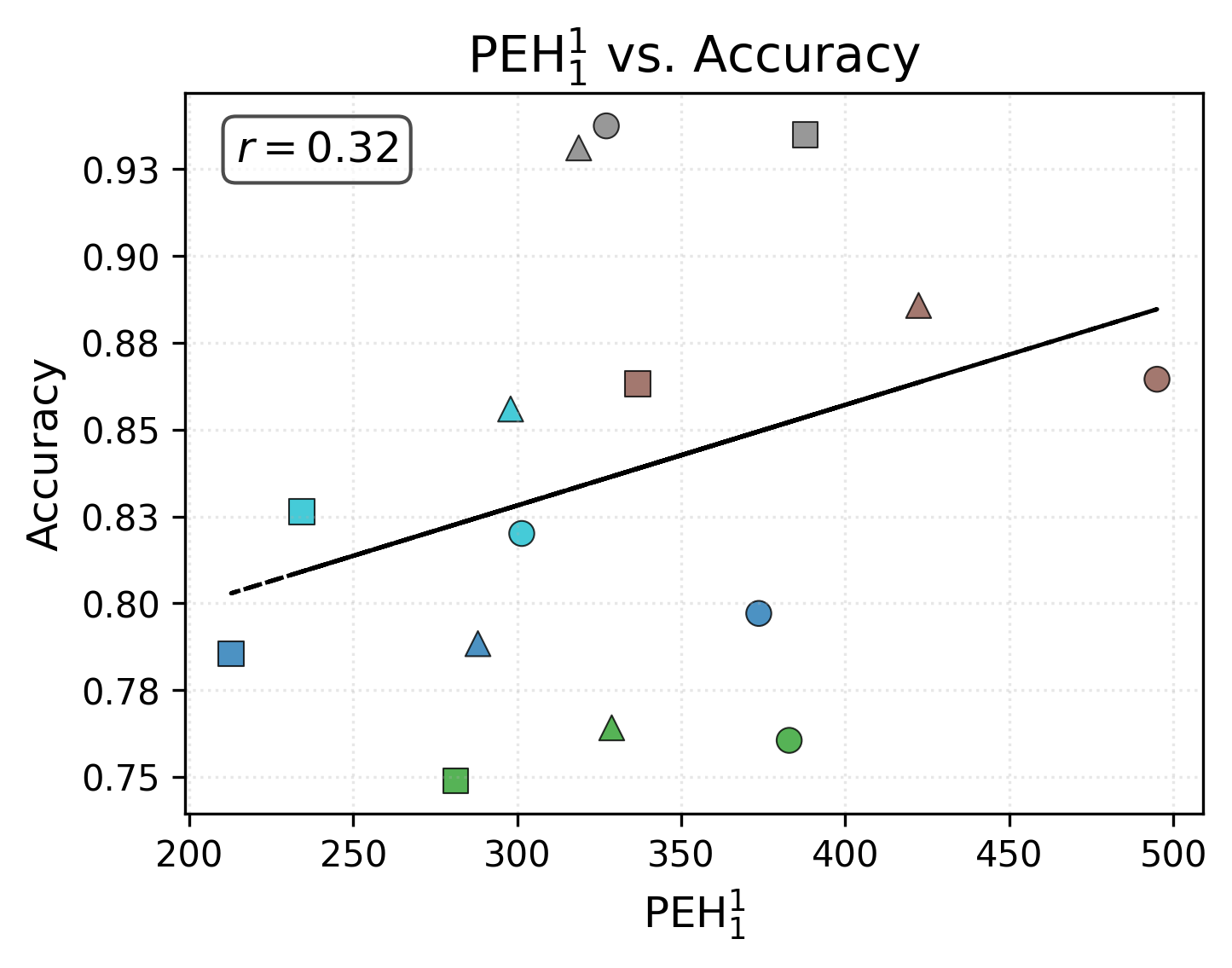}
    \includegraphics[width=0.33\textwidth]{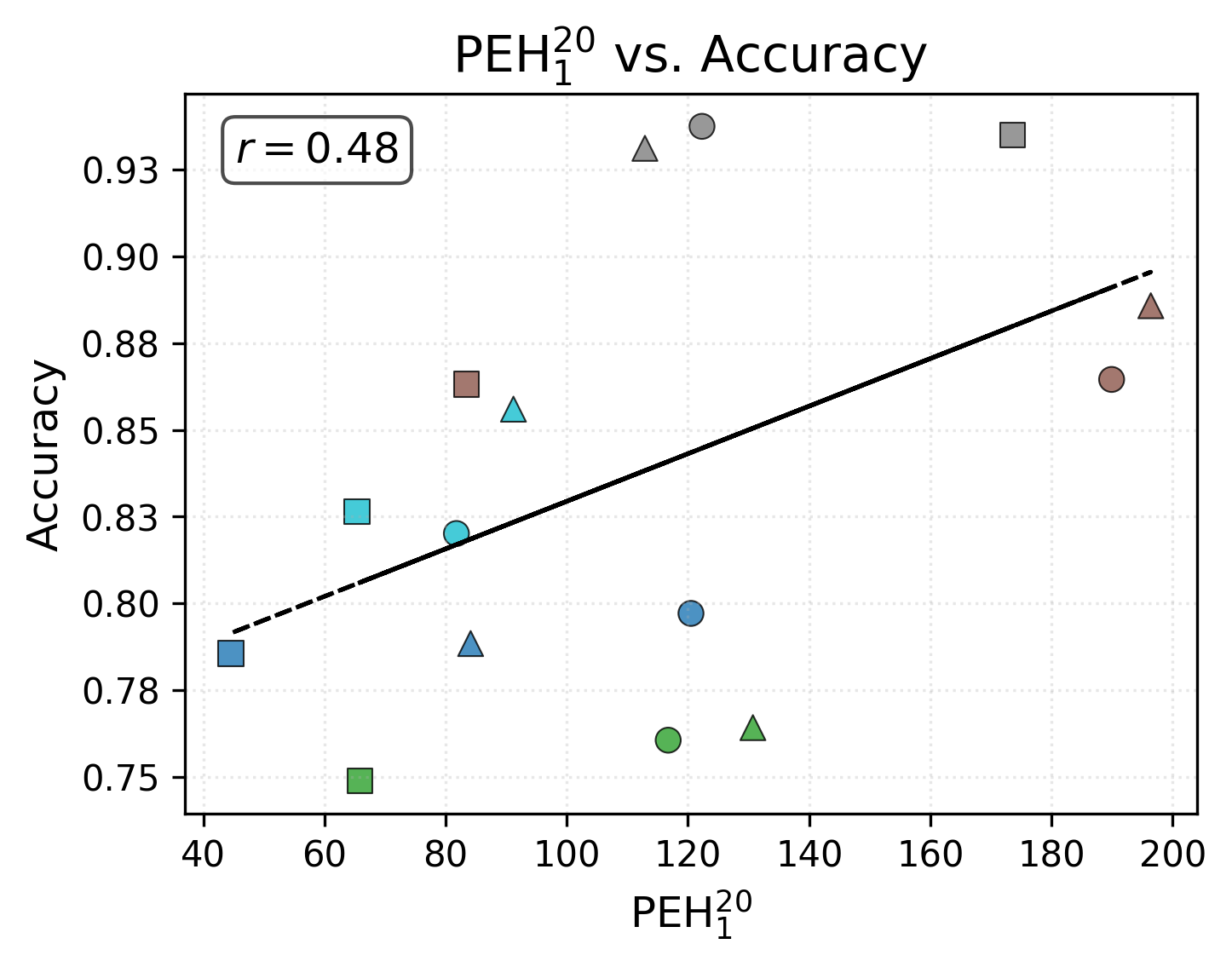}
    \includegraphics[width=0.33\textwidth]{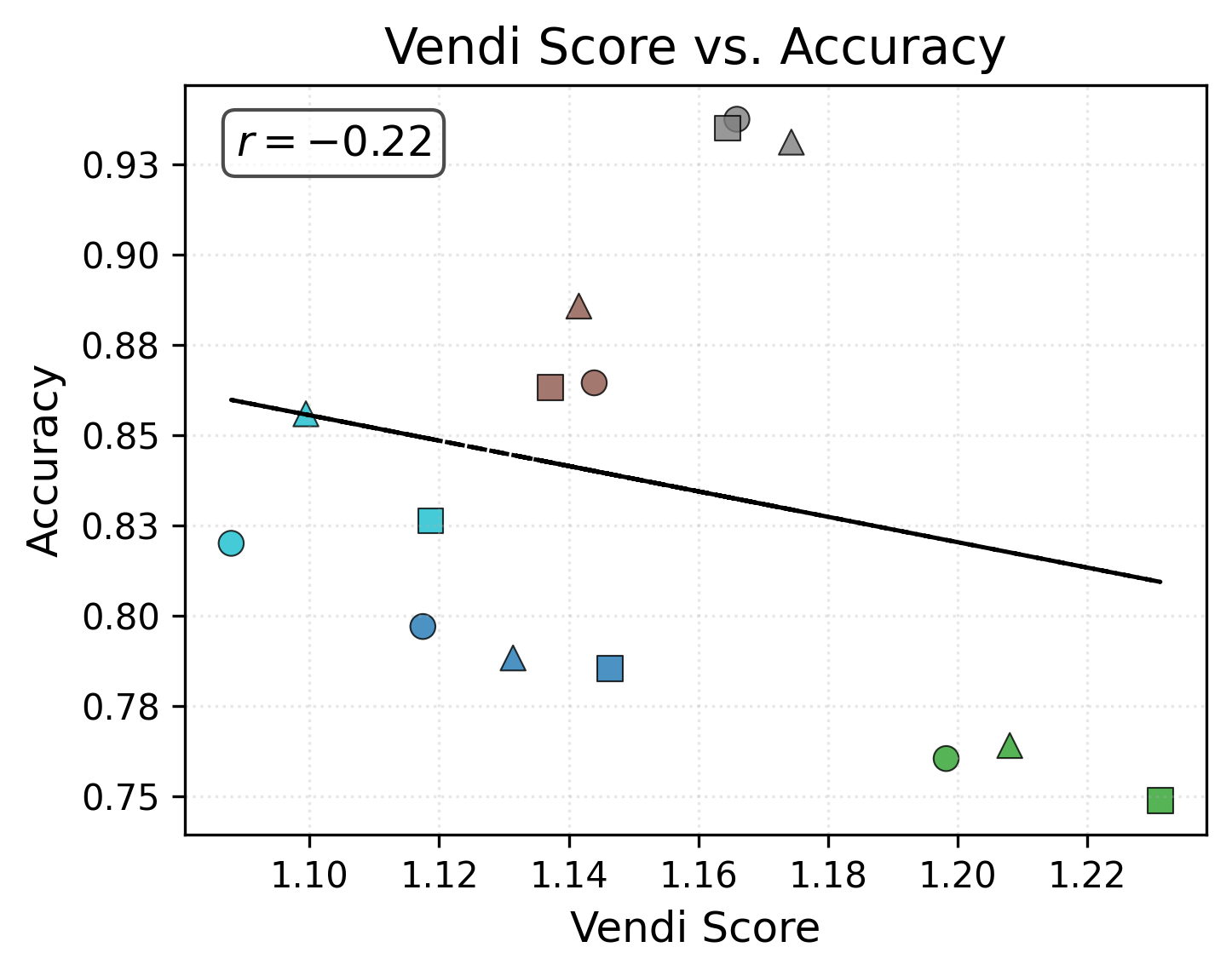}
    \label{fig:h1}
    \end{subfigure}
  \caption{Model Accuracy vs. PH-based Diversity Measures}
  \label{fig:sec4}
\end{figure}

\textbf{Can the data geometry predict its quality?} 
We define high-quality data as data that enables models trained on it to achieve high accuracy. Figure \ref{fig:sec4} illustrates that PH-based diversity measures are positively correlated with model accuracy; i.e., greater diversity in both $H_0$ (connected components) and $H_1$ (loops) features tends to improve model performance. By contrast, the Vendi Score exhibits a negative trend, with higher values associated with lower accuracy. This contrast highlights that geometry-aware indicators, such as PH-based measures, can serve as reliable predictors of data quality, whereas distributional, entropy-based measures, like Vendi Score, fail to do so (the last plot in Figure \ref{fig:sec4}). We also observe a negative relationship between minimum $H_0$ values and the standard deviation of model accuracy. This suggests that greater geometric diversity not only boosts accuracy but also stabilizes model training, an effect detailed in the Appendix (Figure \ref{fig:std}).

\textbf{Which data geometry characteristics are most desirable for an effective model training?} 
Table \ref{tab:catergory} shows that the topological features of different subsets strongly influence model performance. The \emph{closest} subset has extremely high $H_0$-based Hill numbers, indicating fragmented clusters with many clusters of similar persistence, and high $H_1$-based Hill numbers with short lifetimes, reflecting noisy and unstable loops. It yields slightly better average accuracy than the farthest subset, but suffers from the largest variance, highlighting instability caused by redundancy and noise. The \emph{farthest} subset shows both low $H_0$- and $H_1$-based Hill numbers, resulting in sparse, fragile clusters and minimal loop diversity. The scattered points provide poor coverage of the data manifold, producing the lowest accuracy, though with slightly better stability than the \emph{closest} subset. The \emph{random} subset strikes a balance. Moderate $H_0$-based Hill numbers correspond to well-separated, non-redundant clusters, while moderate $H_1$-based Hill numbers reflect stable loops. This balanced geometric profile produces the best accuracy and lowest standard deviation, representing the most desirable structural diversity for training.


\begin{table}[h]
\caption{Relationship Between PH-based Diversity Measures and Average Accuracy by Category}
\label{tab:catergory}
\centering
\begin{tabular}{lllll}
\cmidrule(r){1-5}
Subset & Accuracy (avg $\pm$ std) & $H_0$ Measure & $H_1$ Measure & Vendi Score \\
\midrule
Closest & 0.836 $\pm$ 0.021 &
\begin{tabular}[t]{@{}l@{}}
$\mathrm{PEH}_0^{1}$:489 \\
$\mathrm{PEH}_0^{20}$:347  \\
$H_0$ min: 0.0215
\end{tabular} &
\begin{tabular}[t]{@{}l@{}}
$\mathrm{PEH}_1^{1}$:376 \\
$\mathrm{PEH}_1^{20}$:126  \\
$H_1$ mean: 0.0025
\end{tabular} &
1.143 \\
\midrule
Farthest & 0.832 $\pm$ 0.014 &
\begin{tabular}[t]{@{}l@{}}
$\mathrm{PEH}_0^{1}$:478 \\
$\mathrm{PEH}_0^{20}$:244  \\
$H_0$ min: 0.0191
\end{tabular} &
\begin{tabular}[t]{@{}l@{}}
$\mathrm{PEH}_1^{1}$:291 \\
$\mathrm{PEH}_1^{20}$:86  \\
$H_1$ mean: 0.0029
\end{tabular} &
1.160 \\
\midrule
Random & 0.845 $\pm$ 0.013 &
\begin{tabular}[t]{@{}l@{}}
$\mathrm{PEH}_0^{1}$:485 \\
$\mathrm{PEH}_0^{20}$:287  \\
$H_0$ min: 0.0234
\end{tabular} &
\begin{tabular}[t]{@{}l@{}}
$\mathrm{PEH}_1^{1}$:331 \\
$\mathrm{PEH}_1^{20}$:123  \\
$H_1$ mean: 0.0028
\end{tabular} &
1.151 \\
\bottomrule
\end{tabular}
\end{table}

\textbf{Takeways.} A high-quality dataset should exhibit well-separated clusters ($H_0$) and contain some stable loops ($H_1$), while avoiding the extremes of redundancy (too many overlapping data points) or sparsity (scattered, fragile structures). Random sampling often achieves this balance,  but more deliberate strategies for data augmentation can explicitly target high $H_0$ minimum values (ensuring separation) combined with moderate $H_1$ mean lifetimes (capturing stable geometric features). We further observe that a relatively smaller training dataset (6\%–19\% of the original dataset) can achieve 91\%–98.6\% of the accuracy reported using the full dataset during fine-tuning \cite{ba-etal-2024-fill}. This finding underscores the critical role of data selection in model performance and highlights that more data is not always better -- what matters is the right structural diversity.

\section{Conclusion}

In this work, we demonstrated that the training data geometry, captured through persistent homology, is closely linked to model performance. Traditional entropy-based diversity metrics alone prove insufficient for predicting training data quality, whereas PH-based diversity measures offer clear advantages by effectively quantifying the structural richness of a dataset. This persistent homology–based method provides insight by separating noise-induced fragmentation from meaningful structural richness in the data. Our analysis of different data subsets shows that well-balanced clusters ($H_0$) combined with stable loops ($H_1$) yield the best accuracy and lowest variability, showing the importance of structural diversity. These insights pave the way for more principled strategies in constructing training datasets, data augmentation, and synthetic data generation. Future research could explore how topological features can be directly leveraged to guide robust model training, improving generalization while reducing dependence on large-scale data.





\bibliographystyle{plainnat}
\bibliography{refs}

\appendix

\section{Technical Appendices and Supplementary Material}
\label{appedix}

\subsection{Proof of Diversity Axioms for PH-Based Measures}

A diversity measure derived from Persistent Homology (PH) is defined as a summary statistic of the persistence lifetimes generated from a dataset's Vietoris-Rips filtration. We prove that such a measure satisfies the key principles of effective size, the twin property, multi-scale analysis, and symmetry. $\mathrm{Div}(X)$ denotes as PH-based Diversity Measure in general.

\begin{itemize}
    \item \textbf{Effective Size}: In a fixed-cardinality dataset, diversity increases when the data points spread out, and decreases when points concentrate or fully coincide, reaching a minimum when all points are identical. For a dataset $X = \{x_1, \dots, x_n\}$ where $x_i = x_j$ for all $i,j$, $\mathrm{Div}(X)$ attains its theoretical minimum value; for a dataset $X_s$ containing distinct points, $\mathrm{Div}(X_s) \geq \mathrm{Div}(X)$.  

    \begin{proof}
    
    \emph{Minimum Diversity (Collapsed Data).}  
    If $x_1 = x_2 = \dots = x_n$, then $D$ is the zero matrix. In the Vietoris--Rips filtration, all points form a single connected component at $\epsilon=0$. No loops ($H_1$ features) appear. Persistence intervals thus have zero or infinite length, degenerating to a trivial diagram. Any summary statistic (e.g., total persistence, entropy, Hill number) computed from this single lifetime yields its minimum possible value, correctly reflecting a minimal diversity or an effective size of one. \\[1ex] 

    \emph{Higher Diversity (Separated Data).}  
    Conversely, if the dataset $X_s$ consists of well-separated points, as $\epsilon$ increases, components merge, creating multiple $H_0$ features with non-zero lifetimes. Moreover, geometric arrangements can generate robust higher-dimensional features, such as loops ($H_1$), that persist across a wide range of scales. The resulting persistence diagram is richer and has a varied distribution of lifetimes. Summary statistics applied to this richer distribution yield a higher value, reflecting the greater effective size and topological complexity of the data.\\[1ex] 

    Thus, the PH-based measure maps collapsed, redundant data to low diversity and structurally rich, separated data to high diversity.
    
    \end{proof}

    \item \textbf{Twin Property}: Duplicating a data point does not change the measured diversity. Let $X$ be a dataset and let $x_i \in X$. For the set $X' = X \cup \{x_n\}$ where $x_n = x_i$, the diversity is unchanged:
    \[
    \mathrm{Div}(X') = \mathrm{Div}(X).
    \]

    \begin{proof}
    By definition, the distance between the twin points is $d(x_i,x_n)=0$.  
    For any other point $x_j \in X$, the distance from the duplicate is identical to the distance from the original: $d(x_n,x_j)=d(x_i,x_j)$. \\[1ex] 
    
    In the Vietoris--Rips filtration, $x_n$ forms a $0$-distance edge with $x_i$ and never contributes a feature with nonzero persistence (its lifetime is $\ell=0$, because the connected components corresponding to $x_i$ and $x_n$ are born at $\epsilon=0$ and merge immediately, generating a persistence interval of $(0,0)$). All other interpoint distances are unchanged, so the persistence diagram of nonzero intervals is invariant. Hence, $\mathrm{Div}(X') = \mathrm{Div}(X).$
    
    \end{proof}

    \item \textbf{Multi-Scale}: The measure accounts for geometric/topological features across the full range of distance scales, capturing both local and global structure. The diversity measure $\mathrm{Div}(X)$ is a function of the entire multiset of persistence lifetimes that integrates information from all scales in the filtration. 

    \begin{proof}
    
    Persistent homology tracks connected components, loops, and higher-dimensional holes across all filtration scales $\epsilon \in [0, \max d(x_i, x_j)]$.
    Long persistence intervals correspond to large-scale, global topological features, while short intervals capture local or noise-induced structures.
    By tuning the filtration bounds—using a smaller $\epsilon_{\max}$ to focus on fine-scale neighborhoods and local diversity, or a larger $\epsilon_{\min}$ to filter out small-scale noise and emphasize global structure—one can selectively highlight different geometric aspects of the data.
    The resulting diversity measure $\mathrm{Div}(X)$ then summarizes the complete distribution of persistence lifetimes, thereby integrating information across both local and global geometric scales.
    \\[1ex] 
    
    Moreover, parameter choices (e.g.\ Hill number order $q$) can adjust sensitivity to rare vs.\ dominant features. For example, $q>1$ gives more weight to long-lived (global) features and $q<1$ gives more weight to short-lived (local) features. Thus, $\mathrm{Div}(X)$  is multi-scale by construction.  

    \end{proof}

    \item \textbf{Symmetry}: The diversity measure is invariant under permuting (re‐ordering) the data points. Let $X=(x_1,\dots,x_n)$ be an ordered sequence of points and let $\pi$ be any permutation of $\{1,\dots,n\}$. For the permuted sequence $X_\pi=(x_{\pi(1)},\dots,x_{\pi(n)})$, we have
    \[
    \mathrm{Div}(X_\pi) = \mathrm{Div}(X).
    \]
    
    \begin{proof}
    The PH pipeline begins with the pairwise distance matrix $D$, where $D_{ij}=d(x_i,x_j)$. Let $X_\pi$ be the reordered dataset. The distance matrix $D_\pi$ for the permuted data has entries $(D_\pi)_{ij}=d(x_{\pi(i)},x_{\pi(j)})$. 
    Importantly, the set of all unique pairwise distances
    \[
    \{d(x_i,x_j)\}_{1 \leq i < j \leq n}
    \]
    is unchanged for both $X$ and $X_\pi$. The construction of the Vietoris--Rips filtration depends only on these distances. Hence, the persistence diagrams and lifetimes $\{l_i\}$ are identical. Therefore, any diversity measure computed from these lifetimes is invariant under permutation of the data.
    \end{proof}
    
\end{itemize}

\subsection{Correlation Between PH-based Diversity Measure and the Standard Deviation of Model Accuracy}

 \begin{figure}[htb]
    \centering
    \includegraphics[width=0.5\linewidth]{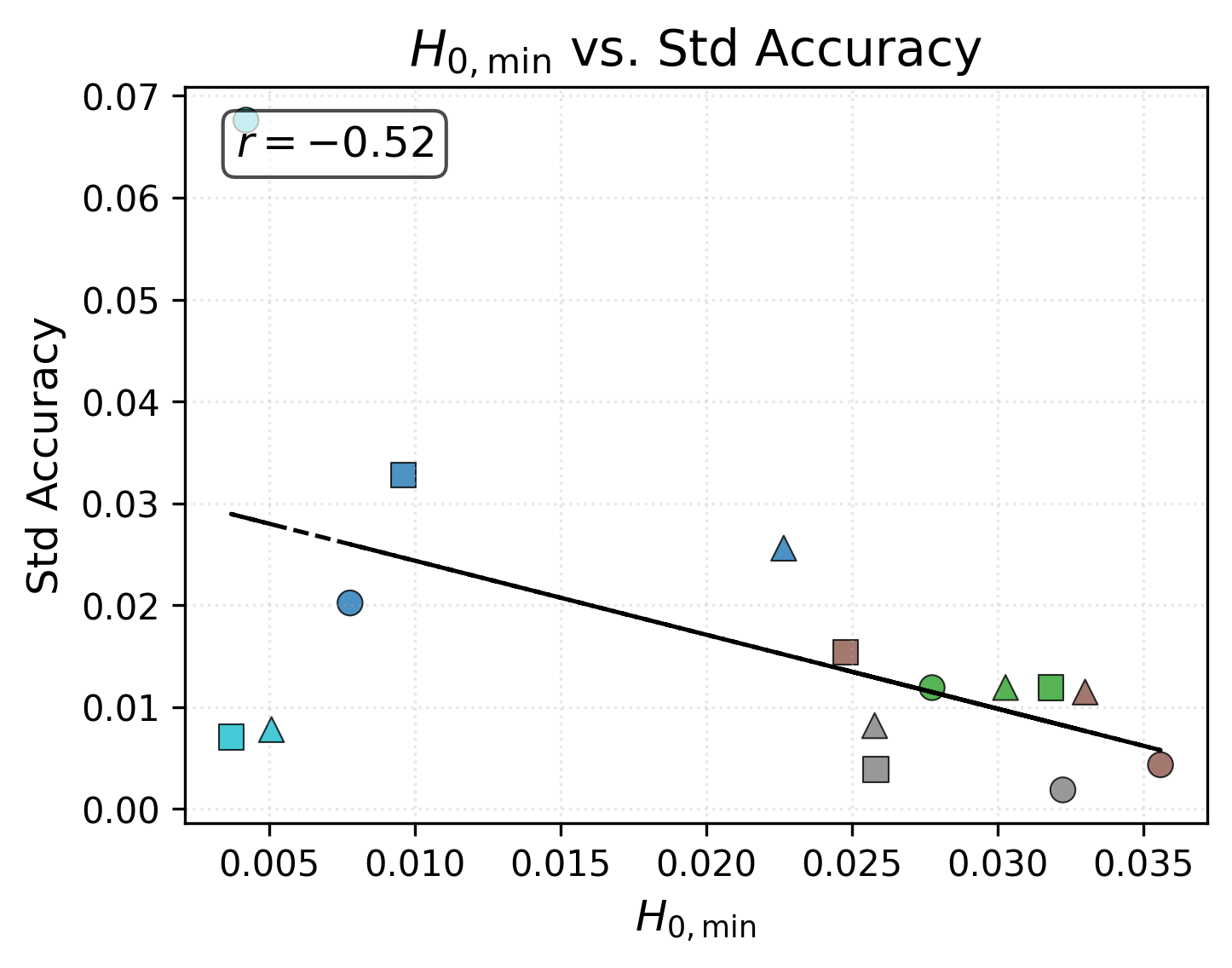}
    \caption{ Minimum $H_0$ values vs. the standard
deviation of model accuracy. A negative correlation between the minimum $H_0$ values and the standard deviation of model accuracy indicates that greater separation between clusters improves the stability of model training.}
    \label{fig:std}
\end{figure}

\end{document}